\pdfoutput=1

\documentclass[11pt,a4paper]{article}
\usepackage[hyperref]{ACL2023}
\usepackage{times}
\usepackage{inconsolata}
\usepackage{latexsym}
\usepackage{danudefs}
\usepackage{algorithmic}
\usepackage{algorithm}
\usepackage{color}
\usepackage{booktabs}
\usepackage{xspace}
\usepackage{url}
\usepackage{microtype}

\usepackage{subcaption}
\usepackage{multirow}
\usepackage{amsmath}
\usepackage{pifont}% http://ctan.org/pkg/pifont

\newcommand{\xmark}{\ding{55}}%

\usepackage[english]{babel}
\usepackage{hyperref}
\addto\captionsenglish{%
}
\addto\extrasenglish{%
}

\usepackage{todonotes}

\makeatletter
\if@todonotes@disabled

\else

\fi
\makeatother

\title{Unsupervised Semantic Variation Prediction \\using the Distribution of Sibling Embeddings}

\author{Taichi Aida \\
  Tokyo Metropolitan University \\
  {\tt aida-taichi@ed.tmu.ac.jp}
  \And Danushka Bollegala \\
  Amazon, University of Liverpool \\
  {\tt danushka@liverpool.ac.uk}}

\date{}

\begin{document}
\maketitle
\begin{abstract}
Languages are dynamic entities, where the meanings associated with words constantly change with time.
Detecting the semantic variation of words is an important task for various NLP applications that must make time-sensitive predictions.
Existing work on semantic variation prediction have predominantly focused on comparing some form of an averaged contextualised representation of a target word computed from a given corpus.
However, some of the previously associated meanings of a target word can become obsolete over time (e.g. meaning of \emph{gay} as \emph{happy}), while novel usages of existing words are observed (e.g. meaning of \emph{cell} as a mobile phone).
We argue that mean representations alone cannot accurately capture such semantic variations and propose a method that uses the entire cohort of the contextualised embeddings of the target word, which we refer to as the \emph{sibling distribution}.
Experimental results on SemEval-2020 Task 1 benchmark dataset for semantic variation prediction show that our method outperforms prior work that consider only the mean embeddings, and is comparable to the current state-of-the-art. 
Moreover, a qualitative analysis shows that our method detects important semantic changes in words that are not captured by the existing methods.
\footnote{Source code is available at \url{https://github.com/a1da4/svp-gauss} .}
%\footnote{An anonymised version of the source code to reproduce our evaluations is submitted to the ACL-2023 review system, and will be publicly released upon paper acceptance.}
\end{abstract}

\section{Introduction}
\label{sec:intro}

\begin{figure}[!t]
    \begin{minipage}{\linewidth}
        \centering
        \includegraphics[width=60mm]{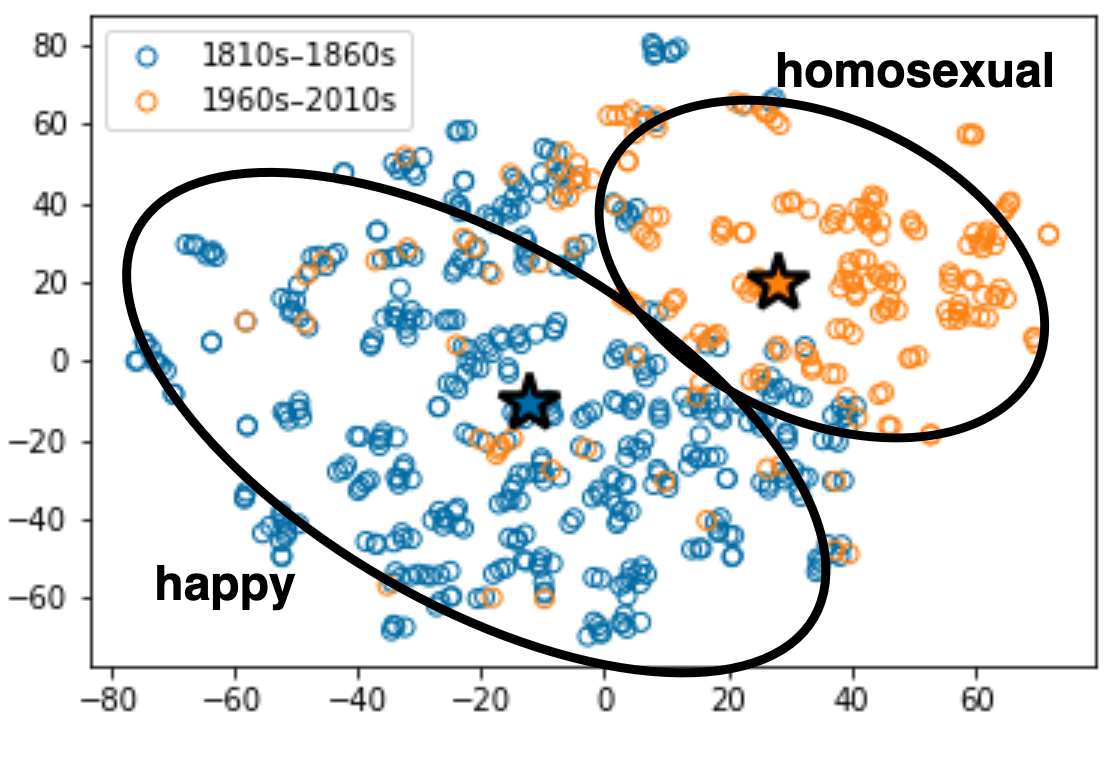}
        \subcaption{gay}\label{fig:gay}
    \end{minipage}\\
    \begin{minipage}{\linewidth}
        \centering
        \includegraphics[width=60mm]{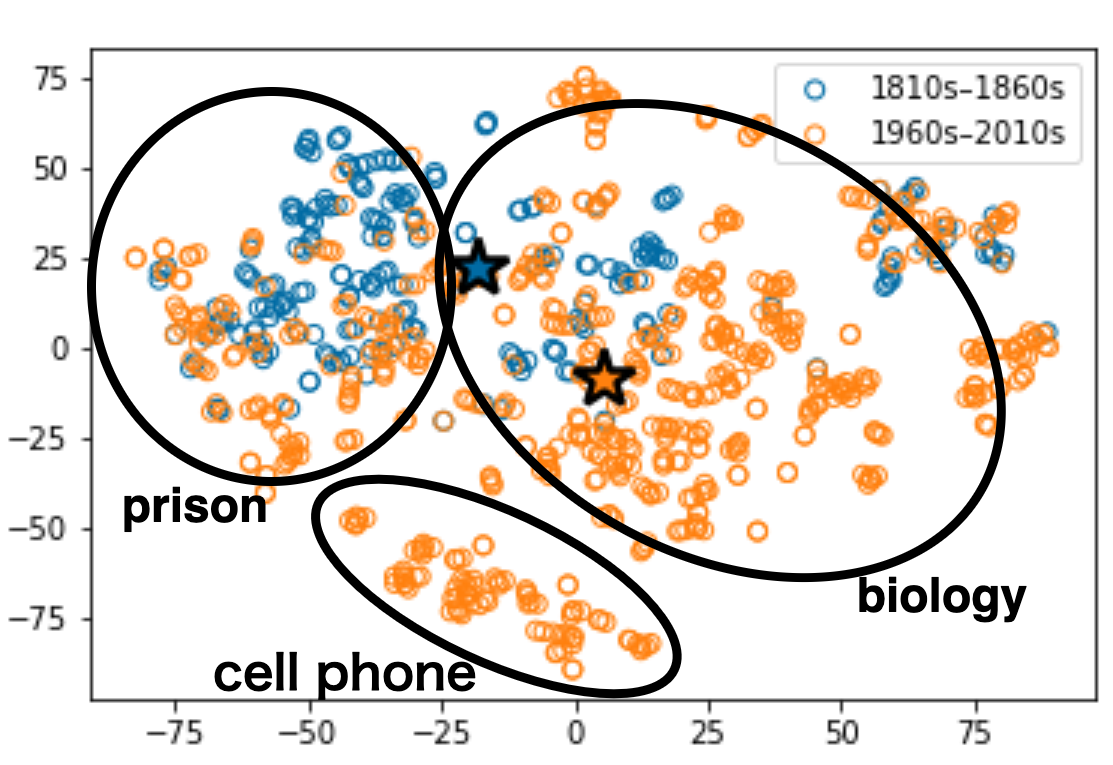}
        \subcaption{cell}\label{fig:cell}
    \end{minipage}
    \caption{t-SNE projections of BERT token vectors (dotted) in two time periods and the average vector (starred) for each period. (a) the word \textit{gay} has lost its original meaning related to \textit{happy} and is now used to mean \textit{homosexual}, resulting in a significant shift in its distribution. (b) the word \textit{cell} is now also used to mean \textit{cell phone}, while retaining the meaning of \textit{prison} or \textit{biology}, widening the distribution but not significantly changing the mean vector.}\label{fig:outline}
    \vspace{-5mm}
\end{figure}

The meaning of words evolves over time, and even in everyday life, technological innovations and cultural aspects can cause a word to have a different meaning than in the past.
For example, the meaning of the word \textit{gay} has completely changed from \textit{happy} to \textit{homosexual} (\autoref{fig:gay}), and \textit{cell} has added \textit{cell phone} to its previous meanings of \textit{prison} and \textit{biology} (\autoref{fig:cell}).
In the semantic change detection task, the goal is to detect the words whose meanings have changed across time-specific corpora~\cite{kutuzov-etal-2018-diachronic, tahmasebia-etal-2021-survey}.

As illustrated in \autoref{fig:outline}, we can identify two types of semantic changes associated with words -- (a) the word \textit{gay} obtains a new meaning by \textbf{replacing} its past meaning (\autoref{fig:gay}), whereas (b)
the word \textit{cell} obtains a new meaning, while \textbf{preserving} its past meanings (\autoref{fig:cell}).
On the other hand, much prior work have resort to a scheme where they first individually represent the meaning of a target word in a given time-specific corpora  using a single embedding, such as the mean of the non-contextualised~\cite{kim-etal-2014-temporal, kulkarni-etal-2015-statistically, hamilton-etal-2016-diachronic, yao-etal-2018-dynamic, dubossarsky-etal-2019-time, aida-etal-2021-comprehensive} or
contextualised~\cite{martinc-etal-2020-leveraging, beck-2020-diasense, kutuzov-giulianelli-2020-uio, rosin-etal-2022-time, rosin-radinsky-2022-temporal} embeddings of the target word taken over all of its occurring contexts in the corpus.
Next, various distance measures are used to compare those embeddings to quantify the semantic variation of the target word across corpora.
However, as seen from \autoref{fig:outline}, using the mean embedding of a target word alone for predicting semantic variations of words can be misleading especially when the variance of the embedding distribution is large.

To address the above-mentioned limitations, we use the distribution of contextualised embeddings of a target word $w$ in all of its occurrence contexts $\cS(w)$ in a given corpus, which we refer to as the \emph{sibling distribution}~\cite{zhou-etal-2022-problems} of $w$.
We then approximate the sibling distribution of a word using a multivariate Gaussian, which has shown to accurately capture the uncertainty in word embedding spaces~\cite{vilnis-and-mccallum-2015-word,iwamoto-yukawa-2020-rijp,yuksel-etal-2021-semantic}.
We can then use a broad range of distance and divergence measures defined over Gaussian distributions to quantify the semantic variation of a target word across multiple time-specific corpora.

Experimental results on SemEval-2020 Task 1 benchmark dataset show that our proposed method outperforms several prior methods, and achieves comparable performance to the current state-of-the-art (SoTA)~\cite{rosin-radinsky-2022-temporal}.
More importantly, our proposal to model both the mean and variance of sibling embeddings consistently outperforms methods that use only the mean contextualised embedding from the same Masked Language Model (MLM)~\cite{rosin-radinsky-2022-temporal}.
Moreover, for computational convenience, prior work had assumed the covariance matrix of sibling embeddings to be diagonal ~\cite{iwamoto-yukawa-2020-rijp, yuksel-etal-2021-semantic}, but we show that further performance improvements can be obtained by using the full covariance matrix.

\section{Related Work}
Historically, the diachronic semantic changes of words have been studied by linguists~\cite{tahmasebia-etal-2021-survey}, which has also received much attention lately within the NLP community.
Automatic detection of words whose meanings change over time has provided important insights for diverse fields such as linguistics, lexicology, sociology, and information retrieval (IR)~\cite{traugott-dasher-2001-regularity, cook-stevenson-2010-automatically, michel-etal-2011-quantitative, kutuzov-etal-2018-diachronic}.
For example, in IR one must know the seasonal association of keywords used in user queries to provide relevant results pertaining to a particular time period.
Moreover, it has been shown that the performance of publicly available pretrained foundation models~\cite{FM} declines over time when applied to emerging data~\cite{loureiro-etal-2022-timelms, lazaridou-etal-2021-mind} because they are trained using a static snapshot.
\citet{Su-etal-2022-improving} showed that the temporal generalisation of foundation models is closely related to their ability to detect semantic variations of words.

Semantic change detection is modelled in the literature as an unsupervised task of detecting words whose meanings change between two given time-specific corpora~\cite{kutuzov-etal-2018-diachronic, tahmasebia-etal-2021-survey}.
In recent years, several shared tasks have been held~\cite{schlechtweg-etal-2020-semeval, basile-etal-2020-diacr, kutuzov-etal-2021-rushifteval}, where participants are required to predict the degree or presence of semantic changes for a given target word between two given corpora sampled from different time periods.
For this purpose, much prior work have used non-contextualised or contextualised word embeddings to represent the meaning of the target word in each corpus.
Unlike non-contextualised word embeddings, which represent a word by the same vector in all of its contexts, contextualised word embeddings represent the same target word with different vectors in different contexts.
Various methods have been proposed to map vector spaces from different time periods, such as initialisation~\cite{kim-etal-2014-temporal}, alignment~\cite{kulkarni-etal-2015-statistically, hamilton-etal-2016-diachronic}, and joint learning~\cite{yao-etal-2018-dynamic, dubossarsky-etal-2019-time, aida-etal-2021-comprehensive}.

The existing methods that have been proposed for the semantic variation detection of words can be broadly categorised into two groups: (a) methods that compare word/context clusters~\cite{hu-etal-2019-diachronic, giulianelli-etal-2020-analysing, montariol-etal-2021-scalable}, and (b) methods that compare embeddings of the target words computed from different corpora sampled at different time periods~\cite{martinc-etal-2020-leveraging, beck-2020-diasense, kutuzov-giulianelli-2020-uio, rosin-etal-2022-time}.
Recently, it has been reported that adding time-specific attention mechanisms~\cite{rosin-radinsky-2022-temporal} achieves SoTA performance.
However, this model requires additional training of the entire MLM including the time-specific mechanisms, which is computationally costly for large-scale MLMs.

Despite the recent success of using word embeddings for the semantic change detection task, many of these methods struggle to detect meaning changes of words which have a wide range of usages because they use only the mean embedding to represent a target word~\cite{kutuzov-etal-2022-contextualized}.
Although methods that use point estimates in the embedding space, such as using non-contextualised word embeddings or comparing the average of contextualised word embeddings, are able to detect semantic variations that result in a loss of a prior meaning (e.g. \textit{gay} in \autoref{fig:gay}), they are inadequate when detecting semantic variations due to novel usages of words, while preserving their former meanings  (e.g. \textit{cell} in \autoref{fig:cell}).

To alleviate this problem, some studies have used Gaussian Embeddings~\cite{vilnis-and-mccallum-2015-word} for semantic change detection~\cite{iwamoto-yukawa-2020-rijp, yuksel-etal-2021-semantic}.
They used the mean and the diagonal approximation of the covariance matrix computed using non-contextualised word embeddings.
However, as argued previously, contextualised embeddings provide useful clues regarding the meaning of a word as used in a context.
Therefore, in our proposed method, we consider the entire cohort of contextualised word embeddings of a target word taken across all of its occurring contexts (i.e. siblings) obtained from an MLM.
As confirmed later by the evaluations presented in \autoref{sec:sota}, our proposed method consistently outperforms the methods proposed by \citet{iwamoto-yukawa-2020-rijp} and \citet{yuksel-etal-2021-semantic} that use non-contextualised embeddings.

\section{Semantic Variation Prediction}

Let us consider a target word $w$ that occurs in two given corpora $C_1$ and $C_2$.
For example, $C_1$ and $C_2$ could have been sampled at two distinct time slots, respectively $T_1$ and $T_2$, reflecting any \emph{temporal} semantic variations of words, or alternatively sampled at similar periods in time but from distinct domains (e.g. \emph{biology} vs. \emph{law}) expressing semantic variations of words due to the differences in the \emph{domains}.
Our goal in this paper is to propose a method that can accurately predict whether $w$ is used in the same meaning in both $C_1$ and $C_2$ (i.e. $w$ is semantically invariant across the two corpora) or otherwise (i.e. its meaning is different in the two corpora).
Although we consider two corpora in the subsequent description for simplicity of the disposition, our proposed method can be easily extended to measure the semantic variation of a word over multiple corpora.

According to the distributional hypothesis~\cite{Firth:1957}, the context in which a word occurs provides useful clues regarding its meaning.
Contextualised word embeddings such as the ones produced by MLMs have shown to concisely and accurately encode contextual information related to a target word in a given context.
For example, \newcite{yi-zhou-2021-learning} showed that contextualised word embeddings can be used to induce word-sense embeddings that represent the distinct senses of an ambiguous word with different vectors.
Inspired by such prior work using contextualised word embeddings as a proxy for accessing contextual information related to a target word, we propose a method to detect the semantic variations of a target word using its multiple occurrences in a corpus.

% describe the set of contexts.
% introduce the concept of sibling and cite Ethirajah for the first use
% define the set of sibling emebddings
% approximate it as a Gaussian. Give computational resons for this approximation
% The Gaussian approximation (Sibling Distribution) can be used as-is for computing some metrics such as KL, Jeffrey's. We will take samples from the Sibling distributions to compute the other (instance-based) metrics.
% The metric itself is used as the `semantic variation score`
% Spearman can be computed against the human-ratings then.

To describe our proposed method in detail, let us denote the set of contexts containing $w$ in corpus $C_i$ by $\cS(w,C_i)$.
The scope of the context of $w$ could be limited to a predefined fixed token window or extended to the entire sentence containing $w$ as we do in our experiments.
Let us denote the contextualised (token) embedding of $w$ in a context $s \in \cS(w, C_i)$ produced by an MLM $M$ by $\vec{f}_M(w,s) \in \R^d$, where $d$ is the dimensionality of the token embeddings produced by $M$.
Following the terminology introduced by \newcite{zhou-etal-2022-problems}, we refer to type embedding $\vec{f}_M(w,s)$ as the \emph{sibling} embeddings of $w$ in context $s$.
The number of siblings of $w$ in $C_i$ is denoted by $N^w_i = |\cS(w,C_i)|$.
Moreover, let the set of sibling embeddings of $w$ created from its occurrences in $C_i$ to be $\cD(w,C_i) = \{\vec{f}_M(w,s) | s \in \cS(w,C_i)\}$.
As we later see, the distribution of sibling embeddings of a word $w$ encodes information about the usage of $w$ in a corpus, which is useful for predicting any semantic variations of $w$ across different corpora.

We can obtain a context-independent embedding, $\vec{\mu}^w_i \in \R^d$ for $w$ by averaging all of its sibling embeddings over the contexts as given by \eqref{eq:avg}.
\begin{align}
    \label{eq:avg}
    \vec{\mu}^w_i = \frac{1}{N^w_i} \sum_{s \in \cS(w, C_i)} \vec{f}_M(w,s)
\end{align}
Although much prior work has used $\vec{\mu}^w_i$ as a proxy for the usage of $w$ in $C_i$ for numerous tasks such as studying the properties of contextualised embeddings~\cite{ethayarajh-2019-contextual} and predicting semantic variation of words~\cite{martinc-etal-2020-leveraging, beck-2020-diasense, kutuzov-giulianelli-2020-uio, rosin-etal-2022-time, rosin-radinsky-2022-temporal}, the mean of the sibling embedding distribution is insensitive to the rare yet important usages of the target word.
In particular, when the sibling embedding distribution is not uniformly distributed around its mean, the mean embedding can be misleading as a representation of the distribution.
To overcome this limitation, in addition to $\vec{\mu}^w_i$, we also use the co-variance matrix $\mat{V}^w_i \in \R^{d \times d}$ computed from the sibling embedding distribution of $w$ as defined by \eqref{eq:cov}.
\begin{align}
    \label{eq:cov}
    \mat{V}^w_i = \frac{1}{N^w_i (N^w_i - 1)} \!\! \sum_{s \in \cS(w, C_i)} \!\!\!\! \vec{f}_M(w,s) \vec{f}_M(w,s)\T
\end{align}

We approximate the distribution of sibling embeddings of $w$ using a Gaussian, $\cN(\vec{\mu}^w_i, \mat{V}^w_i)$ with mean and variance given respectively by \eqref{eq:avg} and \eqref{eq:cov}.
Gaussian distribution is the maximum entropy distribution over the real values given a finite mean and covariance and no further information~\cite{Jaynes:2003}.
Moreover, by approximating the sibling distribution as a Gaussian, we can use a broad range of distance and divergence measures for quantifying the semantic variation of $w$ across corpora.
% Add explanation about sibling embedding matrix rank here.
% BERT は文の情報を保存 → 頻度とrankに相関 → rank にも情報が保存されている（？）　→ 我々の手法を後押し
In the field of information theory, MLMs have been shown to store the information of a given sentence in a vector~\cite{pimentel-etal-2020-information}.
There is a strong correlation between the word frequency $N^w_i$ and the rank of its covariance matrix $\mat{V}^w_i$ (\autoref{fig:freq_rank} in \autoref{sec:appendix_freq_rank}), which indicates that covariance matrix also retains important information regarding sibling embedding distribution.
This observation further supports our proposal to represent target words by $\vec{\mu}^w_i$ and $\mat{V}^w_i$.

\subsection{Quantifying Semantic Variations}

Given a target word $w$, following the method described above, we represent $w$ in $C_1$ and $C_2$ respectively by the two Gaussian distributions  $\cN(\vec{\mu}^w_1, \mat{V}^w_1)$ and  $\cN(\vec{\mu}^w_2, \mat{V}^w_2)$.
We can then compute a \emph{semantic variation score} for $w$ that indicates how likely the meaning of $w$ has changed from $C_1$ to $C_2$ by using different distance (or divergence) measures to quantify the differences between two Gaussians. For this purpose, we use two types of measures.

\noindent\textbf{Divergence measures} quantify the divergence between two distributions. We use two divergence measures in our experiments: Kullback-Liebler (KL) divergence and Jeffrey's divergence.
Given that we approximate sibling distribution of $w$ in a corpus by a Gaussian, we can analytically compute both KL and Jeffery's divergence measures using $\vec{\mu}^w_1, \vec{\mu}^w_2, \mat{V}^w_1$ and $\mat{V}^w_2$ in closed-form formulas (\autoref{sec:appendix_divergence}).

\noindent\textbf{Distance measures} are defined between two points in the sibling embedding space. We use the seven distance measures: Bray-Curtis, Canberra, Chebyshev, City Block, Correlation, Cosine, and Euclidean.
The definitions of the distance measures used in this paper are provided in \autoref{sec:appendix_distance}.
Given a distance measure $\psi(\vec{w}_1, \vec{w}_2)$ that takes two $d$-dimensional sibling embeddings of $w$, each computed from contexts selected respectively from $C_1$ and $C_2$ and returns a nonzero real number indicating the distance between $\vec{w}_1$ and $\vec{w}_2$, we compute the semantic variation score, ${\rm score}(w)$, of $w$ between $C_1$ and $C_2$ as the average distance over all pairwise comparisons between the sibling embeddings as given by \eqref{eq:dist}.
\begin{align}
    \label{eq:dist}
    {\rm score}(w) = \frac{1}{N^w_1 N^w_2}\sum_{\substack{\vec{w}_1 \in \cD(w,C_1) \\ \vec{w}_2 \in \cD(w,C_2)}} \psi(\vec{w}_1, \vec{w}_2)
\end{align}
The number of occurrences of some target words $w$ can be significantly different between $C_1$ and $C_2$, which can make the computation of \eqref{eq:dist} biased towards the corpus with more contexts for $w$.
To overcome this issue, instead of using sibling embeddings of $w$ computed from actual occurrence contexts of $w$, we sample equal numbers of sibling embeddings from $\cN(\vec{\mu}^w_1, \mat{V}^w_1)$ and  $\cN(\vec{\mu}^w_2, \mat{V}^w_2)$.
Samples can be drawn efficiently from a multidimensional Gaussian by first drawing samples from a standard normal distribution (i.e. with zero mean and unit variance) and subsequently applying a affine transformation parametrised by the $\vec{\mu}^w_i$ and $\mat{V}^w_i$ of the associated sibling distribution.

\section{Experiments}
\label{sec:exp}

\subsection{Data and Metric}
We use the SemEval-2020 Task 1 English dataset\footnote{It is licensed under a Creative Commons Attribution 4.0 International License.}~\cite{schlechtweg-etal-2020-semeval} to evaluate the performance in detecting words whose meanings change between time periods.
This task includes two subtasks, classification and ranking.
In the classification task, the words in the evaluation set must be classified as to whether they have semantically changed over time or otherwise. 
Classification accuracy is used as the evaluation metric for this task. 
On the other hand, in the ranking task, the words in the evaluation set must be sorted according to the degree of semantic change.
Spearman's rank correlation coefficient between the human-rated gold scores and the induced ranking scores is used as the evaluation metric for this task.
In this study, the evaluation is conducted on the ranking task using English data.
We do not perform the classification task because no validation set is available for tuning a classification threshold.

Statistics of the data used in our experiments are %summarised
in \autoref{tab:dataset}.
This data includes two corpora from different centuries extracted from CCOHA~\cite{alatrash-etal-2020-ccoha}.
Let us denote the early 1800s and late 1900s to early 2000s corpora respectively by $C_1$ and $C_2$.
The test set has 37 target words that are selected for indicating whether they have undergone a semantic change between the two time periods.
%These words are annotated for the presence and degree of semantic change.
%These words are manually annotated indicating whether their meaning has changed over time and if so for the degree of their semantic change.
These words are annotated indicating whether their meaning has changed over time and the degree of their semantic change.

\begin{table}[t]
    \centering
    \begin{tabular}{l|rrr} \toprule
        Time Period & \#Sentences & \#Tokens & \#Types \\ \midrule
        1810s--1860s & 254k & 6.5M & 87k \\
        1960s--2010s & 354k & 6.7M & 150k \\ \bottomrule
    \end{tabular}
    \caption{Statistics of the SemEval-2020 Task 1 English dataset~\cite{schlechtweg-etal-2020-semeval}.}
    \label{tab:dataset}
\end{table}

\subsection{Setup}
We use two types of BERT-base models as the MLM in our experiments: a publicly available pretrained model\footnote{\url{https://huggingface.co/bert-base-uncased}} (MLM$_{\textit{pre}}$) and a fine-tuned model (MLM$_{\textit{temp}}$) from MLM$_{\textit{pre}}$~\cite{rosin-etal-2022-time}.
The base model consists of 12 layers, which we use in two different configurations: (a) we use the last layer (MLM$_{\textit{pre} | \textit{temp}, \textrm{last}}$), and (b) the mean-pool over the last four layers (MLM$_{\textit{pre} | \textit{temp}, \textrm{four}}$), which has shown good performance across languages following \citet{laicher-etal-2021-explaining}.
\citet{rosin-radinsky-2022-temporal} recommend using the mean pooling over all (12) hidden layers.
However, we found no statistically significant differences between the mean-pool over all layers vs. the last four layers in our preliminary experiments.

In the prediction of the degree of semantic change for a given word, the set of sibling embeddings for each time period $\cD(w,C_1)$ and $\cD(w,C_2)$ is acquired from all occurrences in each corpus using the MLM described above, and the distributions across time periods $\cN(\vec{\mu}^w_1, \mat{V}^w_1)$ and $\cN(\vec{\mu}^w_2, \mat{V}^w_2)$ are compared. 
For calculating the seven distance measures, we sample 1,000 sibling embeddings from each sibling distribution.
We use the covariance matrix of the sibling embedding, which defines the distribution, only for the diagonal components (\textit{diag}(\textit{cov})) in the divergence measures,\footnote{In the above two divergences, it is necessary to calculate the inverse of the covariance matrix, but in the case of full components, it is often impossible to calculate the inverse matrix because it is not regular.} and both diagonal and full components (\textit{full}(\textit{cov})) in the distance measures.
Previous studies assume that the covariance matrix is diagonal (\textit{diag}(\textit{cov}))~\cite{iwamoto-yukawa-2020-rijp, yuksel-etal-2021-semantic}.
This assumption increases computational efficiency compared to \textit{full}(\textit{cov}), at the expense of loosing information on the non-diagonal elements.
In our settings, representation of a sibling distribution $\cN(\vec{\mu}^w_i, \mat{V}^w_i)$ in \textit{diag}(\textit{cov}) or \textit{full}(\textit{cov}) requires $2d$ or $d(1+d)$ parameters, respectively.

\begin{table*}[t!]
    \centering
    \begin{tabular}{l|rrrr} \toprule
         & \multicolumn{4}{c}{Model} \\ 
        Measure & MLM$_{\textit{pre}, \textrm{last}}$ & MLM$_{\textit{pre}, \textrm{four}}$ & MLM$_{\textit{temp}, \textrm{last}}$ & MLM$_{\textit{temp}, \textrm{four}}$ \\ \midrule
        KL($C_1||C_2$) & 0.075 & 0.130 & 0.414 & 0.431\\
        KL($C_2||C_1$) & 0.100 & 0.117 & 0.361 & 0.411\\
        Jeff($C_1||C_2$) & 0.090 & 0.129 & 0.391 & 0.409\\
        Bray-Curtis & \textbf{0.217} & 0.241 & 0.464 & 0.480\\
        Canberra & 0.192 & 0.251 & 0.455 & \textbf{0.517}\\
        Chebyshev & 0.154 & 0.166 & \textbf{0.517} & 0.478\\
        City Block & 0.198 & 0.140 & 0.461 & 0.459\\
        Correlation & 0.191 & 0.266 & 0.480 & 0.463\\
        Cosine & 0.190 & \textbf{0.270} & 0.478 & 0.480\\
        Euclidean & 0.198 & 0.249 & 0.473 & 0.474\\\bottomrule
    \end{tabular}
    \caption{Results of two divergences and seven distance functions under various MLM conditions with the proposed method using \textit{diag}(\textit{cov}). The best performance in each MLM condition is shown in \textbf{bold}. $C_1$ and $C_2$ refer to the early 1800s and late 1900s to early 2000s corpora, respectively. We report two types of KL divergence because of its asymmetric nature. Unlike KL divergence, Jeffrey's divergence is symmetric, and we report just one result.}
    \label{tab:result_intrinsic_diag}
\end{table*}

\begin{table*}[t!]
    \centering
    \begin{tabular}{l|rrrr} \toprule
         & \multicolumn{4}{c}{Model} \\ 
        Measure & MLM$_{\textit{pre}, \textrm{last}}$ & MLM$_{\textit{pre}, \textrm{four}}$ & MLM$_{\textit{temp}, \textrm{last}}$ & MLM$_{\textit{temp}, \textrm{four}}$ \\ \midrule
        Bray-Curtis & \textbf{0.219} & 0.263 & 0.460 & 0.467\\
        Canberra & 0.195 & 0.246 & 0.502 & \textbf{0.489}\\
        Chebyshev & 0.145 & 0.132 & \textbf{0.529} & 0.451\\
        City Block & 0.192 & 0.248 & 0.414 & 0.452\\
        Correlation & 0.181 & \textbf{0.286} & 0.481 & 0.468\\
        Cosine & 0.189 & 0.272 & 0.479 & 0.454\\
        Euclidean & 0.204 & 0.231 & 0.454 & 0.457\\\bottomrule
    \end{tabular}
    \caption{Results of two divergences and seven distance functions under various MLM conditions with the proposed method using \textit{full}(\textit{cov}). The best performance in each MLM condition is shown in \textbf{bold}.}
    \label{tab:result_intrinsic_full}
\end{table*}

\subsection{Result}
We show the results of the proposed method under various conditions in \autoref{tab:result_intrinsic_diag} and \autoref{tab:result_intrinsic_full}.
As reported in previous studies~\cite{rosin-etal-2022-time, rosin-radinsky-2022-temporal}, we find that the fine-tuned model (MLM$_{\textit{temp}}$) achieves high performance in all settings.
Moreover, for the hidden layers, we have confirmed that our method, by using the last four layers (MLM$_{\textit{pre} | \textit{temp}, \textrm{four}}$), yields even higher correlations than using only the last layer (MLM$_{\textit{pre} | \textit{temp}, \textrm{last}}$).

\paragraph{Prediction measures.} Our method allows us to try a variety of measures. 
In the \textit{diag}(\textit{cov}) setting, we try two divergences and seven distance measures.
Comparing within divergence measures, \autoref{tab:result_intrinsic_diag} shows that $\textrm{KL}(C_1||C_2)$ achieves high performance in all MLM conditions.
This result means that many existing words acquire novel meanings.
On the other hand, comparing the distance measures, we find that Canberra and Chebyshev outperform the commonly used cosine distance in MLM$_{\textit{temp}}$ (\autoref{tab:result_intrinsic_diag} and \autoref{tab:result_intrinsic_full}). 
Since the cosine distance makes underestimations in MLMs \cite{zhou-etal-2022-problems}, this result suggests that it is better to calculate the absolute distance per dimension as in Canberra and Chebyshev.

\paragraph{Components of the covariance matrices.} When applying the distance measures, the vectors can be extracted from the full or diagonal covariance matrix. 
From \autoref{tab:result_intrinsic_full} we see that using all components of the covariance matrix (\textit{full}(\textit{cov})) further improves performance obtaining a correlation coefficient of $0.529$ (MLM$_{\textit{temp}, \textrm{last}}$, \textit{full}(\textit{cov}), Chebyshev). 
Previous studies had assumed that the covariance matrix is diagonal for computational convenience~\cite{iwamoto-yukawa-2020-rijp, yuksel-etal-2021-semantic}.
However, as our results show, further performance improvements can be obtained by considering all components of the covariance matrix.
Here onwards, we will refer to the best setting (i.e. MLM$_{\textit{temp}, \textrm{last}}$, \textit{full}(\textit{cov}), Chebyshev) as the \textbf{Proposed} method.

\subsection{Comparisons against Prior work}
\label{sec:sota}

In this section, we compare our proposed method against related prior work.
We do not re-implement or re-run those methods, but instead compare using the published results from the original papers.

\begin{description}

    \item \textbf{Word2Gauss}$_{\textrm{light}}$ \cite{iwamoto-yukawa-2020-rijp}: They apply Gaussian Embeddings~\cite{vilnis-and-mccallum-2015-word} based architecture in each time period. For each word, they define a computationally lightweight Gaussian embedding as follows: the mean vector is the vector of the word2vec learned by the initialization method~\cite{kim-etal-2014-temporal}, and the covariance matrix is the diagonal matrix, uniformly weighted by frequency. They calculate the KL divergence of the Gaussian embeddings for the semantic variation prediction.
    
    \item \textbf{Word2Gauss} \cite{yuksel-etal-2021-semantic}: They apply pure Gaussian Embeddings~\cite{vilnis-and-mccallum-2015-word}. For a given word, the mean vector and the covariance matrix of the Gaussian Embedding are trained using the inner-product with the positive examples and the KL divergence with the negative examples. For computational convenience and to reduce the number of parameters, they use a diagonal covariance matrix. 
    After training separate word embedding models for each time period, the mean vectors are aligned between time periods using a rotation matrix~\cite{hamilton-etal-2016-diachronic}, and predictions are made using cosine distance or Jeffrey's divergence.
    They have reported the cosine distance as the best metric. %We use the cosine distance, which was reported as the best metric in their paper.
    
    \item \textbf{MLM}$_{\textit{temp}}$ \cite{rosin-etal-2022-time}: They fine-tuned the published BERT model to specific time periods. To adapt to specific time periods, they insert a special token indicating the time period at the beginning of the sentence in the target corpus, and fine-tuned on the corpora available for each time period. 
    They use two measures for prediction: (a) the distance between the predicted probability of the target word in the sentence at each time period, and
    (b) the cosine distance of the average token vector at each time period. 
    Their results report that the cosine distance is the best metric (\textbf{MLM}$_{\textit{temp}}$,~Cosine).
    %We use the cosine distance, which was reported as the best metric in their paper.
    However, \newcite{kutuzov-giulianelli-2020-uio} have shown that the average pairwise cosine distance \eqref{eq:dist} is better than the cosine distance between average sibling embeddings. 
    Based on this result, we only run this setting that {MLM}$_{\textit{temp}}$ model with the average pairwise cosine distance (\textbf{MLM}$_{\textit{temp}}$,~APD).
    
    \item \textbf{MLM}$_{\textit{pre}}$ \textbf{w/ Temp. Att.} \cite{rosin-radinsky-2022-temporal}: They propose a time-specific attention mechanism to adapt MLMs to specific time periods. They add time-specific vectors and an attention weight matrix to the published BERT as trainable parameters and perform additional training on the target corpora. 
    During prediction, they use the cosine distance following \citet{rosin-etal-2022-time}.
    
    \item \textbf{MLM}$_{\textit{temp}}$ \textbf{w/ Temp. Att.} \cite{rosin-radinsky-2022-temporal}: %It is considered as the current SoTA model for semantic variation detection.
    It is the combination of the above two methods (\textbf{MLM}$_{\textit{temp}}$ and \textbf{MLM}$_{\textit{pre}}$ \textbf{w/ Temp. Att.}), which is considered as the current SoTA model for semantic variation prediction. 
    They add time-specific special tokens to the beginning of each sentence in the target corpus, and conduct additional training on the publicly available BERT model with the time-specific attention mechanism. 
    They also use the cosine distance as used by \citet{rosin-etal-2022-time}.
\end{description}

\begin{table}[t]
    \centering
    \begin{tabular}{l|r} \toprule
        Model & Spearman \\ \midrule
        Word2Gauss$_{\textrm{light}}$ & 0.358 \\
        Word2Gauss & 0.399 \\
        MLM$_\textit{temp}$, Cosine & 0.467 \\
        MLM$_\textit{temp}$, APD & 0.479 \\
        MLM$_\textit{pre}$ w/ Temp. Att. & 0.520 \\
        MLM$_\textit{temp}$ w/ Temp. Att. & \textbf{0.548} \\
        Proposed & \underline{\textit{0.529}} \\ \bottomrule
    \end{tabular}
    \caption{Comparison against prior work including SoTA. In our method, we report the top three results and all of the cosine distance results. The best performance is shown in \textbf{bold}, and the second best is shown in \underline{\textit{underlined}}.}
    \label{tab:result_sota}
\end{table}

Experimental results are summarised in \autoref{tab:result_sota}.
This result shows that our proposed method achieves the second best performance compared to prior work.
We can see that the contextualised mean embeddings based method (\textbf{MLM}$_\textit{temp}$) outperforms the non-contextualised distribution based methods (\textbf{Word2Gauss}$_\textrm{light}$ and \textbf{Word2Gauss}), and further improvement can be obtained by adding the time-specific attention mechanisms (\textbf{MLM}$_\textit{pre}$ \textbf{w/ Temp. Att.} and \textbf{MLM}$_\textit{temp}$ \textbf{w/ Temp. Att.}).
Moreover, the contextualised distribution based approach (\textbf{Proposed}) can yield performance improvement similar to adding time-specific attention mechanisms.
We will discuss the detailed analyses as follows. 

\paragraph{Comparison within the base model (MLM$_{\textit{temp}}$).} Since our method is based on MLM$_{\textit{temp}}$, we compare performance within MLM$_{\textit{temp}}$.
As in the previous work~\cite{rosin-etal-2022-time}, we discuss the results when using the cosine distance.
\autoref{tab:result_sota} shows that the average pairwise cosine distance (MLM$_\textit{temp}$,~APD) outperforms the cosine distance between average sibling embeddings (MLM$_\textit{temp}$,~Cosine).
Moreover, from \autoref{tab:result_intrinsic_diag} and \autoref{tab:result_intrinsic_full}, we can see that our distribution based method outperforms the previous method using only the mean embeddings (0.467 in \autoref{tab:result_sota}) in most settings (0.478 by MLM$_{\textit{temp}, \textrm{last}}$,~\textit{diag}(\textit{cov}), 0.480 by MLM$_{\textit{temp}, \textrm{four}}$,~\textit{diag}(\textit{cov}), and 0.479 by MLM$_{\textit{temp}, \textrm{last}}$,~\textit{full}(\textit{cov})).
This result indicates the importance of considering not only the mean but also the variance of the sibling embeddings.

\paragraph{Comparison against SoTA.} Although our proposed method and the SoTA \textbf{MLM}$_{\textit{temp}}$ \textbf{w/ Temp. Att.} are based on the same model MLM$_{\textit{temp}}$, their configurations are significantly different. 
Specifically, \textbf{MLM}$_{\textit{temp}}$ \textbf{w/ Temp. Att.} adds a time-specific attention mechanism to the model and learns its parameters with additional training, whereas our proposed method uses only MLM$_{\textit{temp}}$ and thus does \emph{not} require additional parameters or training. 
Although according to \autoref{tab:result_sota}, \textbf{MLM}$_{\textit{temp}}$ \textbf{w/ Temp. Att.} reports a correlation of 0.548 and marginally outperforms the Proposed method, which obtains a correlation of 0.529, we find no statistically significant difference between those two methods.\footnote{To measure the statistical significance, we use the Fisher transformation~\cite{Fisher-1992-statistical}.}

\subsection{Ablation Study}
%We conduct an ablation study to understand the importance of considering the covariance $\mat{V}^w_i$, in addition to the mean $\vec{\mu}^w_i$, for the purpose of detecting semantic variations of words.
We conduct an ablation study to understand the importance of (i) predicting semantic variation with sibling distributions $\cN(\vec{\mu}^w_i, \mat{V}^w_i)$, and (ii) constructing sibling distributions from the mean $\vec{\mu}^w_i$ and covariance $\mat{V}^w_i$ of sibling embeddings.
Based on our best setting \textbf{Proposed} (MLM$_{\textit{temp}, \textrm{last}}$, \textit{full}(\textit{cov}), Chebyshev), we define two variants: (i) predicting semantic variation score using mean vectors $\vec{\mu}^w_1$ and $\vec{\mu}^w_2$ only as previous studies, and (ii) constructing a sibling distribution with the identity matrix $\cN(\vec{\mu}^w_i, \mat{I})$ instead of the covariance matrix $\mat{V}^w_i$.
In the SemEval-2020 Task 1 English evaluation set, the existence of a semantic change (binary judgement) and its degree (continuous judgement) are provided. 
Therefore, due to the limited space, we analyse the top eight semantically changed words with the highest degrees of semantic changes and the bottom eight semantically stable words with the lowest degrees of semantic change.
%Based on our best setting \textbf{Proposed} (MLM$_{\textit{temp}, \textrm{last}}$, \textit{full}(\textit{cov}), Chebyshev), we also consider two variants: (i) constructing a sibling distribution with the identity matrix $\cN(\vec{\mu}^w_i, \mat{I})$ instead of the covariance matrix $\mat{V}^w_i$, and (ii) predicting semantic variation score using mean vectors $\vec{\mu}^w_1$ and $\vec{\mu}^w_2$ only as previous studies.

From \autoref{tab:analysis}, we see that our distribution-based variants ($\mat{V}^w_i = \mat{I}$ and Proposed) eliminate overestimation or underestimation problems in using mean vectors only (w/o $\mat{V}^w_i$).
The variant w/o $\mat{V}^w_i$ correctly detects words \textit{plane} and \textit{graft} that have changed meaning significantly between time periods.
However, this variant also reports underestimation (\textit{stab} and \textit{bit}) and overestimation (\textit{contemplation} and \textit{chairman}) in other words, whose meanings are changed/stable but the mean vectors are changed little/significantly.
This is because it makes predictions based only on the mean of sibling embeddings.
On the other side, the distribution-based variants ($\mat{V}^w_i = \mat{I}$ and Proposed) can appropriately rank semantically changed words ($\Delta = \textrm{\checkmark}$) that have small changes in mean vectors (\textit{stab} and \textit{bit}), and stable words ($\Delta = \textrm{\xmark}$) that have large changes in mean vectors (\textit{contemplation} and \textit{chairman}).\footnote{The distribution-based methods fail to detect highly ambiguous words with distinct word senses (\textit{plane} and \textit{graft}). However, the proposed method approximates the distribution of embeddings for a word using a ``single'' Gaussian. We believe by using a mixture of Gaussian this issue can be resolved.}
Moreover, we find that even with the distribution-based variants, using covariance matrices $\mat{V}^w_i$ computed from sibling embeddings yields even better performance than identity matrices ($\mat{V}^w_i = \mat{I}$). 
This result further verifies our hypothesis that considering the mean and the variance of the sibling embeddings is important for semantic change detection tasks.

\begin{table}[t]
\small
    \centering
    \begin{tabular}{l|p{5mm}p{5mm}|p{6mm}p{10mm}p{8mm}} \toprule
        \multirow{2}{1em}{Word} & \multicolumn{2}{c|}{Gold} & w/o $\mat{V}_i^w$ & $\mat{V}_i^w = \mat{I}$ & Proposed \\
         & rank & $\Delta$ & \multicolumn{1}{c}{rank} & \multicolumn{1}{c}{rank} & \multicolumn{1}{c}{rank} \\ \midrule
        plane & 1 & \checkmark & 3 & 18 & 15 \\
        tip & 2 & \checkmark & 7 & 9 & 7 \\
        prop & 3 & \checkmark & 16 & 1 & 4 \\ 
        graft & 4 & \checkmark & 2 & 36 & 36 \\
        record & 5 & \checkmark & 15 & 12 & 14 \\
        stab & 7 & \checkmark & 31 & 10 & 11 \\
        bit & 9 & \checkmark & 27 & 11 & 9 \\ 
        head & 10 & \checkmark & 23 & 28 & 28 \\
        \midrule
        multitude & 30 & \xmark & 24 & 35 & 35 \\
        savage & 31 & \xmark & 20 & 26 & 26 \\
        contemplation & 32 & \xmark & 1 & 37 & 37 \\ 
        tree & 33 & \xmark & 33 & 31 & 30 \\
        relationship & 34 & \xmark & 26 & 34 & 34 \\
        fiction & 35 & \xmark & 21 & 29 & 29 \\
        chairman & 36 & \xmark & 5 & 32 & 33 \\
        risk & 37 & \xmark & 10 & 19 & 21 \\
        \midrule
        \midrule
        Spearman & \multicolumn{2}{c|}{1.000} & 0.070 & 0.503 & \textbf{0.529} \\
        \bottomrule
    \end{tabular}
    \caption{Ablation study on the top-8 semantically changed ($\Delta=$ \checkmark) words with the highest degree of semantic change and the bottom-8 stable words ($\Delta=$ \xmark) with the lowest degree of semantic change. w/o $\mat{V}_i^w$ predicts using mean vectors $\vec{\mu}_1^w$ and $\vec{\mu}_2^w$ directly, whereas $\mat{V}_i^w = \mat{I}$ samples sibling embeddings from a Gaussian with the identity variance (i.e. $\cN(\vec{\mu}_i^w, \mat{I})$) instead of $\cN(\vec{\mu}_i^w, \mat{V}_i^w)$.}
    \label{tab:analysis}
\end{table}

\section{Conclusion}
% Future work
%% 今後性能の良いLMが登場することでさらなる精度向上が期待できる（Limitation と被る）
%% さらなる柔軟な分布の使用（Gaussian Mixture), さらなる分析？（Temporal Word-in-Context タスク https://aclanthology.org/2022.coling-1.296）→タスクの方は下手に挙げるとつつかれるかも

We proposed a method to detect semantic variations of words using sibling embeddings.
Experimental results on SemEval-2020 Task~1 English dataset show that the proposed method consistently outperforms methods that use only the mean embedding vectors, and reports results comparable to the current SoTA.
Furthermore, a qualitative analysis shows that the proposed method correctly detects semantic variation of words, which are either over/underestimated by the existing methods.

\iffalse
In this paper, we proposed a method comparing the distributions of contextualised word embeddings to predict the semantic variation of words.
Our method considers the mean and variance of the set of contextualised word embeddings (\textit{sibling embeddings}) to alleviate the problem of using point estimation caused in prior work.
We experimented with the semantic variation prediction task, SemEval-2020 Task~1.
Experimental results show that our method outperforms the point estimation based method using the same masked language model, and achieves comparable performance against the current state-of-the-art model which requires additional architectures and training.
In a qualitative analysis, our method properly detects the semantic variation of words which the existing method makes overestimations or underestimations.
\fi

\section{Limitations}
\paragraph{Language-related limitations.}
For the ease of the analysis, we conducted experiments using only the English dataset in this study.
Although our proposed method can be applied to any language, its performance must be evaluated on languages other than English.
For example, the SemEval-2020 Task 1 dataset includes Latin, German, and Swedish language datasets, in addition to English, and can be used for this purpose.
In particular, our proposed method requires only pretrained MLMs and does not require additional training data for the target languages, which makes it easily scalable to many languages.

\paragraph{Availability of MLMs for the target language.}
Experimental results show that the quality of the MLM is an important factor determining the performance of the proposed method.
For example, the proposed method reports good performance with vanilla BERT model in \autoref{tab:result_intrinsic_diag} but further gains in performance can be obtained with the fine-tuned BERT model on masked time stamps.
However, since our method assumes the availability of pretrained MLMs, a problem arises when trying to adapt our method to minor languages where no pretrained MLMs are available.
This limitation could be mitigated to an extent by using multilingual MLMs.
For example, \citet{arefyev-zhikov-2020-bos} demonstrated that satisfactory levels of accuracies can be obtained for semantic change detection by using multilingual MLMs.
%it encourages the use of multilingual MLMs in our method even when there are no publicly-available MLMs in the target language.
Our proposed method can further benefit from the fact that new and larger MLMs are being publicly released for many languages in the NLP community.
%In the future, the performance of our method is expected to improve with the advent of even larger and more diverse language models than are currently available.

\section{Ethical Considerations}
In this paper, we proposed a distribution based method using publicly available MLMs, and evaluated with the SemEval-2020 Task 1 English data. 
Although we have not published any datasets or models, \citet{basta-etal-2019-evaluating} shows that pretrained MLMs encode and even amplify unfair social biases such as gender or racial biases.
Given that we obtain sibling distributions from such potentially socially biased MLMs, we must further evaluate the sensitivity of our method for such undesirable social biases.

\section*{Acknowledgements}
This work was supported by JST, the establishment of university fellowships towards the creation of science technology innovation, Grant Number JPMJFS2139. Danushka Bollegala holds concurrent appointments as a Professor at University of Liverpool and as an Amazon Scholar. This paper describes work performed at the University of Liverpool and is not associated with Amazon.

\bibliography{myrefs.bib}

\begin{thebibliography}{42}
\expandafter\ifx\csname natexlab\endcsname\relax\def\natexlab#1{#1}\fi

\bibitem[{Aida et~al.(2021)Aida, Komachi, Ogiso, Takamura, and
  Mochihashi}]{aida-etal-2021-comprehensive}
Taichi Aida, Mamoru Komachi, Toshinobu Ogiso, Hiroya Takamura, and Daichi
  Mochihashi. 2021.
\newblock \href {https://aclanthology.org/2021.paclic-1.3} {A comprehensive
  analysis of {PMI}-based models for measuring semantic differences}.
\newblock In \emph{Proceedings of the 35th Pacific Asia Conference on Language,
  Information and Computation}, pages 21--31, Shanghai, China. Association for
  Computational Lingustics.

\bibitem[{Alatrash et~al.(2020)Alatrash, Schlechtweg, Kuhn, and Schulte~im
  Walde}]{alatrash-etal-2020-ccoha}
Reem Alatrash, Dominik Schlechtweg, Jonas Kuhn, and Sabine Schulte~im Walde.
  2020.
\newblock \href {https://aclanthology.org/2020.lrec-1.859} {{CCOHA}: Clean
  corpus of historical {A}merican {E}nglish}.
\newblock In \emph{Proceedings of the Twelfth Language Resources and Evaluation
  Conference}, pages 6958--6966, Marseille, France. European Language Resources
  Association.

\bibitem[{Arefyev and Zhikov(2020)}]{arefyev-zhikov-2020-bos}
Nikolay Arefyev and Vasily Zhikov. 2020.
\newblock \href {https://doi.org/10.18653/v1/2020.semeval-1.20} {{BOS} at
  {S}em{E}val-2020 task 1: Word sense induction via lexical substitution for
  lexical semantic change detection}.
\newblock In \emph{Proceedings of the Fourteenth Workshop on Semantic
  Evaluation}, pages 171--179, Barcelona (online). International Committee for
  Computational Linguistics.

\bibitem[{Basile et~al.(2020)Basile, Caputo, Caselli, Cassotti, and
  Varvara}]{basile-etal-2020-diacr}
Pierpaolo Basile, Annalina Caputo, Tommaso Caselli, Pierluigi Cassotti, and
  Rossella Varvara. 2020.
\newblock Diacr-ita @ evalita2020: Overview of the evalita2020 diachronic
  lexical semantics (diacr-ita) task.
\newblock In \emph{Proceedings of the Seventh Evaluation Campaign of Natural
  Language Processing and Speech Tools for Italian. Final Workshop (EVALITA
  2020)}. CEUR Workshop Proceedings (CEUR-WS.org).
\newblock Evaluation Campaign of Natural Language Processing and Speech Tools
  for Italian, EVALITA 2020 ; Conference date: 17-12-2020.

\bibitem[{Basta et~al.(2019)Basta, Costa-juss{\`a}, and
  Casas}]{basta-etal-2019-evaluating}
Christine Basta, Marta~R. Costa-juss{\`a}, and Noe Casas. 2019.
\newblock \href {https://doi.org/10.18653/v1/W19-3805} {Evaluating the
  underlying gender bias in contextualized word embeddings}.
\newblock In \emph{Proceedings of the First Workshop on Gender Bias in Natural
  Language Processing}, pages 33--39, Florence, Italy. Association for
  Computational Linguistics.

\bibitem[{Beck(2020)}]{beck-2020-diasense}
Christin Beck. 2020.
\newblock \href {https://doi.org/10.18653/v1/2020.semeval-1.4} {{D}ia{S}ense at
  {S}em{E}val-2020 task 1: Modeling sense change via pre-trained {BERT}
  embeddings}.
\newblock In \emph{Proceedings of the Fourteenth Workshop on Semantic
  Evaluation}, pages 50--58, Barcelona (online). International Committee for
  Computational Linguistics.

\bibitem[{Bommasani et~al.(2021)Bommasani, Hudson, Adeli, Altman, Arora, von
  Arx, Bernstein, Bohg, Bosselut, Brunskill, Brynjolfsson, Buch, Card,
  Castellon, Chatterji, Chen, Creel, Davis, Demszky, Donahue, Doumbouya,
  Durmus, Ermon, Etchemendy, Ethayarajh, Fei-Fei, Finn, Gale, Gillespie, Goel,
  Goodman, Grossman, Guha, Hashimoto, Henderson, Hewitt, Ho, Hong, Hsu, Huang,
  Icard, Jain, Jurafsky, Kalluri, Karamcheti, Keeling, Khani, Khattab, Koh,
  Krass, Krishna, Kuditipudi, Kumar, Ladhak, Lee, Lee, Leskovec, Levent, Li,
  Li, Ma, Malik, Manning, Mirchandani, Mitchell, Munyikwa, Nair, Narayan,
  Narayanan, Newman, Nie, Niebles, Nilforoshan, Nyarko, Ogut, Orr,
  Papadimitriou, Park, Piech, Portelance, Potts, Raghunathan, Reich, Ren, Rong,
  Roohani, Ruiz, Ryan, R{\'e}, Sadigh, Sagawa, Santhanam, Shih, Srinivasan,
  Tamkin, Taori, Thomas, Tram{\'e}r, Wang, Wang, Wu, Wu, Wu, Xie, Yasunaga,
  You, Zaharia, Zhang, Zhang, Zhang, Zhang, Zheng, Zhou, and Liang}]{FM}
Rishi Bommasani, Drew~A. Hudson, Ehsan Adeli, Russ Altman, Simran Arora, Sydney
  von Arx, Michael~S. Bernstein, Jeannette Bohg, Antoine Bosselut, Emma
  Brunskill, Erik Brynjolfsson, Shyamal Buch, Dallas Card, Rodrigo Castellon,
  Niladri Chatterji, Annie Chen, Kathleen Creel, Jared~Quincy Davis, Dora
  Demszky, Chris Donahue, Moussa Doumbouya, Esin Durmus, Stefano Ermon, John
  Etchemendy, Kawin Ethayarajh, Li~Fei-Fei, Chelsea Finn, Trevor Gale, Lauren
  Gillespie, Karan Goel, Noah Goodman, Shelby Grossman, Neel Guha, Tatsunori
  Hashimoto, Peter Henderson, John Hewitt, Daniel~E. Ho, Jenny Hong, Kyle Hsu,
  Jing Huang, Thomas Icard, Saahil Jain, Dan Jurafsky, Pratyusha Kalluri,
  Siddharth Karamcheti, Geoff Keeling, Fereshte Khani, Omar Khattab, Pang~Wei
  Koh, Mark Krass, Ranjay Krishna, Rohith Kuditipudi, Ananya Kumar, Faisal
  Ladhak, Mina Lee, Tony Lee, Jure Leskovec, Isabelle Levent, Xiang~Lisa Li,
  Xuechen Li, Tengyu Ma, Ali Malik, Christopher~D. Manning, Suvir Mirchandani,
  Eric Mitchell, Zanele Munyikwa, Suraj Nair, Avanika Narayan, Deepak
  Narayanan, Ben Newman, Allen Nie, Juan~Carlos Niebles, Hamed Nilforoshan,
  Julian Nyarko, Giray Ogut, Laurel Orr, Isabel Papadimitriou, Joon~Sung Park,
  Chris Piech, Eva Portelance, Christopher Potts, Aditi Raghunathan, Rob Reich,
  Hongyu Ren, Frieda Rong, Yusuf Roohani, Camilo Ruiz, Jack Ryan, Christopher
  R{\'e}, Dorsa Sadigh, Shiori Sagawa, Keshav Santhanam, Andy Shih, Krishnan
  Srinivasan, Alex Tamkin, Rohan Taori, Armin~W. Thomas, Florian Tram{\'e}r,
  Rose~E. Wang, William Wang, Bohan Wu, Jiajun Wu, Yuhuai Wu, Sang~Michael Xie,
  Michihiro Yasunaga, Jiaxuan You, Matei Zaharia, Michael Zhang, Tianyi Zhang,
  Xikun Zhang, Yuhui Zhang, Lucia Zheng, Kaitlyn Zhou, and Percy Liang. 2021.
\newblock {O}n the {O}pportunities and {R}isks of {F}oundation {M}odels.

\bibitem[{Cook and Stevenson(2010)}]{cook-stevenson-2010-automatically}
Paul Cook and Suzanne Stevenson. 2010.
\newblock \href
  {http://www.lrec-conf.org/proceedings/lrec2010/pdf/657_Paper.pdf}
  {Automatically identifying changes in the semantic orientation of words}.
\newblock In \emph{Proceedings of the Seventh International Conference on
  Language Resources and Evaluation ({LREC}'10)}, Valletta, Malta. European
  Language Resources Association (ELRA).

\bibitem[{Dubossarsky et~al.(2019)Dubossarsky, Hengchen, Tahmasebi, and
  Schlechtweg}]{dubossarsky-etal-2019-time}
Haim Dubossarsky, Simon Hengchen, Nina Tahmasebi, and Dominik Schlechtweg.
  2019.
\newblock \href {https://doi.org/10.18653/v1/P19-1044} {Time-out: Temporal
  referencing for robust modeling of lexical semantic change}.
\newblock In \emph{Proceedings of the 57th Annual Meeting of the Association
  for Computational Linguistics}, pages 457--470, Florence, Italy. Association
  for Computational Linguistics.

\bibitem[{Ethayarajh(2019)}]{ethayarajh-2019-contextual}
Kawin Ethayarajh. 2019.
\newblock How contextual are contextualized word representations? comparing the
  geometry of {BERT}, {ELM}o, and {GPT}-2 embeddings.
\newblock In \emph{Proceedings of the 2019 Conference on Empirical Methods in
  Natural Language Processing and the 9th International Joint Conference on
  Natural Language Processing (EMNLP-IJCNLP)}, pages 55--65, Hong Kong, China.
  Association for Computational Linguistics.

\bibitem[{Firth(1957)}]{Firth:1957}
John~R. Firth. 1957.
\newblock A synopsis of linguistic theory 1930-55.
\newblock \emph{Studies in Linguistic Analysis}, pages 1 -- 32.

\bibitem[{Fisher(1992)}]{Fisher-1992-statistical}
R.~A. Fisher. 1992.
\newblock \href {https://doi.org/10.1007/978-1-4612-4380-9_6}
  {\emph{Statistical Methods for Research Workers}}, pages 66--70. Springer New
  York, New York, NY.

\bibitem[{Giulianelli et~al.(2020)Giulianelli, Del~Tredici, and
  Fern{\'a}ndez}]{giulianelli-etal-2020-analysing}
Mario Giulianelli, Marco Del~Tredici, and Raquel Fern{\'a}ndez. 2020.
\newblock \href {https://doi.org/10.18653/v1/2020.acl-main.365} {Analysing
  lexical semantic change with contextualised word representations}.
\newblock In \emph{Proceedings of the 58th Annual Meeting of the Association
  for Computational Linguistics}, pages 3960--3973, Online. Association for
  Computational Linguistics.

\bibitem[{Hamilton et~al.(2016)Hamilton, Leskovec, and
  Jurafsky}]{hamilton-etal-2016-diachronic}
William~L. Hamilton, Jure Leskovec, and Dan Jurafsky. 2016.
\newblock \href {https://doi.org/10.18653/v1/P16-1141} {Diachronic word
  embeddings reveal statistical laws of semantic change}.
\newblock In \emph{Proceedings of the 54th Annual Meeting of the Association
  for Computational Linguistics (Volume 1: Long Papers)}, pages 1489--1501,
  Berlin, Germany. Association for Computational Linguistics.

\bibitem[{Hu et~al.(2019)Hu, Li, and Liang}]{hu-etal-2019-diachronic}
Renfen Hu, Shen Li, and Shichen Liang. 2019.
\newblock \href {https://doi.org/10.18653/v1/P19-1379} {Diachronic sense
  modeling with deep contextualized word embeddings: An ecological view}.
\newblock In \emph{Proceedings of the 57th Annual Meeting of the Association
  for Computational Linguistics}, pages 3899--3908, Florence, Italy.
  Association for Computational Linguistics.

\bibitem[{Iwamoto and Yukawa(2020)}]{iwamoto-yukawa-2020-rijp}
Ran Iwamoto and Masahiro Yukawa. 2020.
\newblock \href {https://doi.org/10.18653/v1/2020.semeval-1.10} {{RIJP} at
  {S}em{E}val-2020 task 1: {G}aussian-based embeddings for semantic change
  detection}.
\newblock In \emph{Proceedings of the Fourteenth Workshop on Semantic
  Evaluation}, pages 98--104, Barcelona (online). International Committee for
  Computational Linguistics.

\bibitem[{Jaynes(2003)}]{Jaynes:2003}
E.~T. Jaynes. 2003.
\newblock \emph{Probability Theory}.
\newblock Cambridge University Press.

\bibitem[{Kim et~al.(2014)Kim, Chiu, Hanaki, Hegde, and
  Petrov}]{kim-etal-2014-temporal}
Yoon Kim, Yi-I Chiu, Kentaro Hanaki, Darshan Hegde, and Slav Petrov. 2014.
\newblock \href {https://doi.org/10.3115/v1/W14-2517} {Temporal analysis of
  language through neural language models}.
\newblock In \emph{Proceedings of the {ACL} 2014 Workshop on Language
  Technologies and Computational Social Science}, pages 61--65, Baltimore, MD,
  USA. Association for Computational Linguistics.

\bibitem[{Kulkarni et~al.(2015)Kulkarni, Al-Rfou, Perozzi, and
  Skiena}]{kulkarni-etal-2015-statistically}
Vivek Kulkarni, Rami Al-Rfou, Bryan Perozzi, and Steven Skiena. 2015.
\newblock Statistically significant detection of linguistic change.
\newblock In \emph{WWW 2015}, pages 625--635.

\bibitem[{Kutuzov et~al.(2022)Kutuzov, Velldal, and
  Ovrelid}]{kutuzov-etal-2022-contextualized}
Andrei Kutuzov, Erik Velldal, and Lilja Ovrelid. 2022.
\newblock \href {https://doi.org/10.3384/nejlt.2000-1533.2022.3478}
  {Contextualized embeddings for semantic change detection: Lessons learned}.
\newblock \emph{Northern European Journal of Language Technology}, 8.

\bibitem[{Kutuzov and Giulianelli(2020)}]{kutuzov-giulianelli-2020-uio}
Andrey Kutuzov and Mario Giulianelli. 2020.
\newblock \href {https://doi.org/10.18653/v1/2020.semeval-1.14}
  {{U}i{O}-{U}v{A} at {S}em{E}val-2020 task 1: Contextualised embeddings for
  lexical semantic change detection}.
\newblock In \emph{Proceedings of the Fourteenth Workshop on Semantic
  Evaluation}, pages 126--134, Barcelona (online). International Committee for
  Computational Linguistics.

\bibitem[{Kutuzov et~al.(2018)Kutuzov, Ovrelid, Szymanski, and
  Velldal}]{kutuzov-etal-2018-diachronic}
Andrey Kutuzov, Lilja Ovrelid, Terrence Szymanski, and Erik Velldal. 2018.
\newblock \href {https://aclanthology.org/C18-1117} {Diachronic word embeddings
  and semantic shifts: a survey}.
\newblock In \emph{Proceedings of the 27th International Conference on
  Computational Linguistics}, pages 1384--1397, Santa Fe, New Mexico, USA.
  Association for Computational Linguistics.

\bibitem[{Kutuzov and Pivovarova(2021)}]{kutuzov-etal-2021-rushifteval}
Andrey Kutuzov and Lidia Pivovarova. 2021.
\newblock Ru{S}hift{E}val: a shared task on semantic shift detection for
  {R}ussian.
\newblock In \emph{Computational linguistics and intellectual technologies:
  Papers from the annual conference Dialogue}.

\bibitem[{Laicher et~al.(2021)Laicher, Kurtyigit, Schlechtweg, Kuhn, and
  Schulte~im Walde}]{laicher-etal-2021-explaining}
Severin Laicher, Sinan Kurtyigit, Dominik Schlechtweg, Jonas Kuhn, and Sabine
  Schulte~im Walde. 2021.
\newblock \href {https://doi.org/10.18653/v1/2021.eacl-srw.25} {Explaining and
  improving {BERT} performance on lexical semantic change detection}.
\newblock In \emph{Proceedings of the 16th Conference of the European Chapter
  of the Association for Computational Linguistics: Student Research Workshop},
  pages 192--202, Online. Association for Computational Linguistics.

\bibitem[{Lazaridou et~al.(2021)Lazaridou, Kuncoro, Gribovskaya, Agrawal,
  Liska, Terzi, Gimenez, de~Masson~d'Autume, Ko{\v{c}}isk{\'y}, Ruder,
  Yogatama, Cao, Young, and Blunsom}]{lazaridou-etal-2021-mind}
Angeliki Lazaridou, Adhiguna Kuncoro, Elena Gribovskaya, Devang Agrawal, Adam
  Liska, Tayfun Terzi, Mai Gimenez, Cyprien de~Masson~d'Autume, Tom{\'a}{\v{s}}
  Ko{\v{c}}isk{\'y}, Sebastian Ruder, Dani Yogatama, Kris Cao, Susannah Young,
  and Phil Blunsom. 2021.
\newblock \href {https://openreview.net/forum?id=73OmmrCfSyy} {Mind the gap:
  Assessing temporal generalization in neural language models}.
\newblock In \emph{Advances in Neural Information Processing Systems}.

\bibitem[{Loureiro et~al.(2022)Loureiro, Barbieri, Neves, Espinosa~Anke, and
  Camacho-collados}]{loureiro-etal-2022-timelms}
Daniel Loureiro, Francesco Barbieri, Leonardo Neves, Luis Espinosa~Anke, and
  Jose Camacho-collados. 2022.
\newblock \href {https://doi.org/10.18653/v1/2022.acl-demo.25} {{T}ime{LM}s:
  Diachronic language models from {T}witter}.
\newblock In \emph{Proceedings of the 60th Annual Meeting of the Association
  for Computational Linguistics: System Demonstrations}, pages 251--260,
  Dublin, Ireland. Association for Computational Linguistics.

\bibitem[{Martinc et~al.(2020)Martinc, Kralj~Novak, and
  Pollak}]{martinc-etal-2020-leveraging}
Matej Martinc, Petra Kralj~Novak, and Senja Pollak. 2020.
\newblock \href {https://aclanthology.org/2020.lrec-1.592} {Leveraging
  contextual embeddings for detecting diachronic semantic shift}.
\newblock In \emph{Proceedings of the Twelfth Language Resources and Evaluation
  Conference}, pages 4811--4819, Marseille, France. European Language Resources
  Association.

\bibitem[{Michel et~al.(2011)Michel, Shen, Aiden, Veres, Gray, null null,
  Pickett, Hoiberg, Clancy, Norvig, Orwant, Pinker, Nowak, and
  Aiden}]{michel-etal-2011-quantitative}
Jean-Baptiste Michel, Yuan~Kui Shen, Aviva~Presser Aiden, Adrian Veres,
  Matthew~K. Gray, null null, Joseph~P. Pickett, Dale Hoiberg, Dan Clancy,
  Peter Norvig, Jon Orwant, Steven Pinker, Martin~A. Nowak, and Erez~Lieberman
  Aiden. 2011.
\newblock \href {https://doi.org/10.1126/science.1199644} {Quantitative
  analysis of culture using millions of digitized books}.
\newblock \emph{Science}, 331(6014):176--182.

\bibitem[{Montariol et~al.(2021)Montariol, Martinc, and
  Pivovarova}]{montariol-etal-2021-scalable}
Syrielle Montariol, Matej Martinc, and Lidia Pivovarova. 2021.
\newblock \href {https://doi.org/10.18653/v1/2021.naacl-main.369} {Scalable and
  interpretable semantic change detection}.
\newblock In \emph{Proceedings of the 2021 Conference of the North American
  Chapter of the Association for Computational Linguistics: Human Language
  Technologies}, pages 4642--4652, Online. Association for Computational
  Linguistics.

\bibitem[{Pimentel et~al.(2020)Pimentel, Valvoda, Maudslay, Zmigrod, Williams,
  and Cotterell}]{pimentel-etal-2020-information}
Tiago Pimentel, Josef Valvoda, Rowan~Hall Maudslay, Ran Zmigrod, Adina
  Williams, and Ryan Cotterell. 2020.
\newblock \href {https://doi.org/10.18653/v1/2020.acl-main.420}
  {Information-theoretic probing for linguistic structure}.
\newblock In \emph{Proceedings of the 58th Annual Meeting of the Association
  for Computational Linguistics}, pages 4609--4622, Online. Association for
  Computational Linguistics.

\bibitem[{Rosin et~al.(2022)Rosin, Guy, and Radinsky}]{rosin-etal-2022-time}
Guy~D. Rosin, Ido Guy, and Kira Radinsky. 2022.
\newblock \href {https://doi.org/10.1145/3488560.3498529} {Time masking for
  temporal language models}.
\newblock In \emph{Proceedings of the Fifteenth ACM International Conference on
  Web Search and Data Mining}, WSDM '22, pages 833--841, New York, NY, USA.
  Association for Computing Machinery.

\bibitem[{Rosin and Radinsky(2022)}]{rosin-radinsky-2022-temporal}
Guy~D. Rosin and Kira Radinsky. 2022.
\newblock \href {https://doi.org/10.18653/v1/2022.findings-naacl.112} {Temporal
  attention for language models}.
\newblock In \emph{Findings of the Association for Computational Linguistics:
  NAACL 2022}, pages 1498--1508, Seattle, United States. Association for
  Computational Linguistics.

\bibitem[{Schlechtweg et~al.(2020)Schlechtweg, McGillivray, Hengchen,
  Dubossarsky, and Tahmasebi}]{schlechtweg-etal-2020-semeval}
Dominik Schlechtweg, Barbara McGillivray, Simon Hengchen, Haim Dubossarsky, and
  Nina Tahmasebi. 2020.
\newblock \href {https://doi.org/10.18653/v1/2020.semeval-1.1}
  {{S}em{E}val-2020 task 1: Unsupervised lexical semantic change detection}.
\newblock In \emph{Proceedings of the Fourteenth Workshop on Semantic
  Evaluation}, pages 1--23, Barcelona (online). International Committee for
  Computational Linguistics.

\bibitem[{Su et~al.(2022)Su, Tang, Guan, Wu, Zhang, and
  Li}]{Su-etal-2022-improving}
Zhaochen Su, Zecheng Tang, Xinyan Guan, Lijun Wu, Min Zhang, and Juntao Li.
  2022.
\newblock \href {https://aclanthology.org/2022.emnlp-main.428} {Improving
  temporal generalization of pre-trained language models with lexical semantic
  change}.
\newblock In \emph{Proceedings of the 2022 Conference on Empirical Methods in
  Natural Language Processing}, pages 6380--6393, Abu Dhabi, United Arab
  Emirates. Association for Computational Linguistics.

\bibitem[{Tahmasebi et~al.(2021)Tahmasebi, Borina, and
  Jatowtb}]{tahmasebia-etal-2021-survey}
Nina Tahmasebi, Lars Borina, and Adam Jatowtb. 2021.
\newblock Survey of computational approaches to lexical semantic change
  detection.
\newblock \emph{Computational approaches to semantic change}, 6:1.

\bibitem[{Traugott and Dasher(2001)}]{traugott-dasher-2001-regularity}
Elizabeth~Closs Traugott and Richard~B. Dasher. 2001.
\newblock \href {https://doi.org/10.1017/CBO9780511486500.004} {\emph{Prior and
  current work on semantic change}}, Cambridge Studies in Linguistics, page
  51–104. Cambridge University Press.

\bibitem[{Vilnis and McCallum(2015)}]{vilnis-and-mccallum-2015-word}
Luke Vilnis and Andrew McCallum. 2015.
\newblock Word representations via gaussian embedding.
\newblock In \emph{Proceedings of the 3rd International Conference on Learning
  Representations}, San Diego, CA, USA.

\bibitem[{Yao et~al.(2018)Yao, Sun, Ding, Rao, and
  Xiong}]{yao-etal-2018-dynamic}
Zijun Yao, Yifan Sun, Weicong Ding, Nikhil Rao, and Hui Xiong. 2018.
\newblock \href {https://doi.org/10.1145/3159652.3159703} {Dynamic word
  embeddings for evolving semantic discovery}.
\newblock In \emph{WSDM 2018}, page 673–681.

\bibitem[{Yüksel et~al.(2021)Yüksel, Uğurlu, and
  Koç}]{yuksel-etal-2021-semantic}
Arda Yüksel, Berke Uğurlu, and Aykut Koç. 2021.
\newblock \href {https://doi.org/10.1109/TASLP.2021.3120645} {Semantic change
  detection with gaussian word embeddings}.
\newblock \emph{IEEE/ACM Transactions on Audio, Speech, and Language
  Processing}, 29:3349--3361.

\bibitem[{Zhou et~al.(2022)Zhou, Ethayarajh, Card, and
  Jurafsky}]{zhou-etal-2022-problems}
Kaitlyn Zhou, Kawin Ethayarajh, Dallas Card, and Dan Jurafsky. 2022.
\newblock Problems with cosine as a measure of embedding similarity for high
  frequency words.
\newblock In \emph{Proceedings of the 60th Annual Meeting of the Association
  for Computational Linguistics (Volume 2: Short Papers)}, pages 401--423,
  Dublin, Ireland. Association for Computational Linguistics.

\bibitem[{Zhou and Bollegala(2021)}]{yi-zhou-2021-learning}
Yi~Zhou and Danushka Bollegala. 2021.
\newblock Learning sense-specific static embeddings using contextualised word
  embeddings as a proxy.
\newblock In \emph{Proceedings of the 35th Pacific Asia Conference on Language,
  Information and Computation}, pages 493--502, Shanghai, China. Association
  for Computational Lingustics.

\bibitem[{Zhou and Bollegala(2022)}]{zhou-bollegala-2022-curious}
Yi~Zhou and Danushka Bollegala. 2022.
\newblock \href {https://aclanthology.org/2022.findings-emnlp.190} {On the
  curious case of l2 norm of sense embeddings}.
\newblock In \emph{Findings of the Association for Computational Linguistics:
  EMNLP 2022}, pages 2593--2602, Abu Dhabi, United Arab Emirates. Association
  for Computational Linguistics.

\end{thebibliography}
\bibliographystyle{acl_natbib}

\appendix
%\section{Information Storage Capacity of the Covariance Matrix}
\section{Information of Sibling Distribution}\label{sec:appendix_freq_rank}
In the semantic variation prediction, prior work have applied the mean embeddings $\vec{\mu}^w_i$ of sibling distribution $\cD(w,C_i)$ for each word $w$. 
However, since these methods compress multiple vectors of $\cD(w,C_i)$ into a single vector $\vec{\mu}^w_i$, there is a risk of loosing the information contained in each vector~\cite{pimentel-etal-2020-information}.
To discuss the amount of information a sibling distribution holds, we analyse the relationship between the size of a sibling distribution $\cD(w,C_i)$ (word frequency $N^w_i$) and the rank of a covariance matrix $\mat{V}^w_i$ calculated from $\cD(w,C_i)$.

\begin{figure}[h]
    \centering
    \includegraphics[width=80mm]{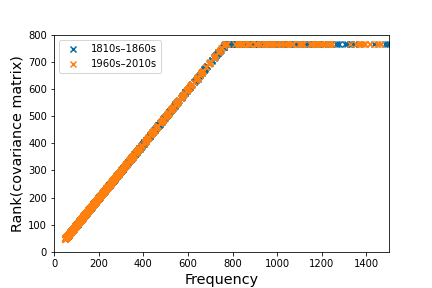}
    \caption{The relationship between the frequency and the rank of the covariance matrix of randomly sampled 1,000 words.}
    \label{fig:freq_rank}
\end{figure}

\autoref{fig:freq_rank} shows the relationship between the frequency of randomly sampled 1,000 words and the rank of their covariance matrices.
For each word, we construct a covariance matrix from sibling embeddings as in \eqref{eq:cov}.
These matrices have $d \times d$ dimensions (BERT base models have $d = 768$ hidden size), and we use their full components (\textit{full}(\textit{cov})) for computing their ranks.
We see that there is a strong correlation between the frequency and the rank of the covariance matrix, and when the frequency exceeds the dimension size, the rank remains constant at the dimensionality of the contextualised embedding space.
This result implies that, upto the dimensionality of the contextualised embedding space, the covariance matrix computed from the sibling distribution $\cD(w,C_i)$, retains information about the individual occurrences of a word. 
Given that contextualised embeddings are often high dimensional (e.g. 768, 1024 etc.) the covariance matrix $\mat{V}^w_i$ computed from the sibling distribution $\cD(w,C_i)$ preserves sufficient information about $w$ for semantic variations related to $w$.

In this analysis, we show that an interesting trend of the word frequency and the rank of covariance matrix.
We speculate that this result may be related to the trend of the sense frequency and the length of sense representation reported in the previous study~\cite{zhou-bollegala-2022-curious}.
However, we leave the investigation of this interesting trend to future research.

\section{List of Divergence Measures}\label{sec:appendix_divergence}
We describe the divergence measures as detailed next.
For simplicity, we denote two Gaussian distributions $\cN(\vec{\mu}^w_1, \mat{V}^w_1)$ and $\cN(\vec{\mu}^w_2, \mat{V}^w_2)$ as $\cN^w_1$ and $\cN^w_2$, respectively.
\begin{description}
    \item \textbf{Kullback-Liebler}
    % 簡単のため、\cN(\vec^w_1, \mat^w_1) を \cN^w_1 とする
    % （もっと適切な表現を探す） を逆行列とする
        \begin{equation}
            \begin{split}
            &\mathrm{KL}(\cN^w_1||\cN^w_2) \\
            =& \frac{1}{2}\Bigl(\mathrm{tr}({\mat{V}^w_2}^{-1}\mat{V}^w_1) - d - \log \frac{\mathrm{det}(\mat{V}^w_1)}{\mathrm{det}(\mat{V}^w_2)}\\
            \ \ \ \ & + (\vec{\mu}^w_2 - \vec{\mu}^w_1)\T{\mat{V}^w_2}^{-1}(\vec{\mu}^w_2 - \vec{\mu}^w_1)\Bigr)
            \end{split}
        \end{equation}
    \item \textbf{Jeffrey's}
        \begin{equation}
            \begin{split}
            &\mathrm{Jeff}(\cN^w_1||\cN^w_2) \\
            =& \frac{1}{2} \mathrm{KL}(\cN^w_1||\cN^w_2) + \frac{1}{2} 
            \mathrm{KL}(\cN^w_2||\cN^w_1) \\
            =& \frac{1}{4}\Bigl(\mathrm{tr}({\mat{V}^w_2}^{-1}\mat{V}^w_1) + \mathrm{tr}({\mat{V}^w_1}^{-1}\mat{V}^w_2) - 2d\\
            \ \ \ \ & + (\vec{\mu}^w_2 - \vec{\mu}^w_1)\T{\mat{V}^w_2}^{-1}(\vec{\mu}^w_2 - \vec{\mu}^w_1) \\
            \ \ \ \ & + (\vec{\mu}^w_1 - \vec{\mu}^w_2)\T{\mat{V}^w_1}^{-1}(\vec{\mu}^w_1 - \vec{\mu}^w_2)\Bigr)
            \end{split}
        \end{equation}
\end{description}

\section{List of Distance Measures}\label{sec:appendix_distance}
We describe the distance measures as detailed next.
$\vec{w}(i)$ denotes the $i$-th value of a word vector $\vec{w}$ and $\overline{\vec{w}}$ denotes a subtracted vector from the average of all dimension values.
\begin{description}
    \item \textbf{Bray-Curtis}
    % \vec(i) は i-th dimension (i \in d) の値を示す
        \begin{equation}
            \psi(\vec{w}_1, \vec{w}_2) = \frac{\sum_{i \in d} |\vec{w}_1(i) - \vec{w}_2(i)|}{\sum_{i \in d} |\vec{w}_1(i) + \vec{w}_2(i)|}
        \end{equation}
    \item \textbf{Canberra}
        \begin{equation}
            \psi(\vec{w}_1, \vec{w}_2) = \sum_{i \in d} \frac{ |\vec{w}_1(i) - \vec{w}_2(i)|}{|\vec{w}_1(i)| + |\vec{w}_2(i)|}
        \end{equation}
    \item \textbf{Chebyshev}
        \begin{equation}
            \psi(\vec{w}_1, \vec{w}_2) = \max_{i} |\vec{w}_1(i) - \vec{w}_2(i)|
        \end{equation}
    \item \textbf{City Block}
        \begin{equation}
            \psi(\vec{w}_1, \vec{w}_2) = \sum_{i \in d} |\vec{w}_1(i) - \vec{w}_2(i)|
        \end{equation}
    \item \textbf{Correlation}
    % \overline{\vec} は各要素から全ての要素の平均を引いたもの
        \begin{equation}
            \psi(\vec{w}_1, \vec{w}_2) = 1 - \frac{\overline{\vec{w}}_1 \cdot \overline{\vec{w}}_2}{||\overline{\vec{w}}_1||_2\ ||\overline{\vec{w}}_2||_2}
        \end{equation}
    \item \textbf{Cosine}
        \begin{equation}
            \psi(\vec{w}_1, \vec{w}_2) = 1 - \frac{\vec{w}_1 \cdot \vec{w}_2}{||\vec{w}_1||_2\ ||\vec{w}_2||_2}
        \end{equation}
    \item \textbf{Euclidean}
        \begin{equation}
            \psi(\vec{w}_1, \vec{w}_2) = ||\vec{w}_1 - \vec{w}_2||_2
        \end{equation}
\end{description}

\end{document}